\pdfoutput=1

\documentclass[conference]{IEEEtran}
\usepackage{times}

\usepackage[numbers]{natbib}
\usepackage{multicol}
\let\labelindent\relax
\usepackage{graphics} 
\usepackage{epsfig} 
\usepackage{mathptmx} 
\usepackage{times} 
\usepackage{amsmath}  
\usepackage{amssymb}  
\usepackage{enumitem}
\usepackage{graphicx}
\usepackage{diagbox}
\usepackage{bm}
\usepackage{makecell}
\usepackage[T1]{fontenc}
\usepackage[latin9]{inputenc}
\usepackage{tikz}
\usepackage{setspace}
\usepackage{url}
\usepackage[bookmarks=true]{hyperref}
\usepackage{flushend}

\usetikzlibrary{shapes,arrows}

\tikzstyle{decision} = [diamond, draw, fill=blue!20, 
    text width=4.5em, text badly centered, node distance=3cm, inner sep=0pt]
\tikzstyle{block} = [rectangle, draw, fill=blue!20, 
    text width=5em, text centered, rounded corners, minimum height=4em]
\tikzstyle{line} = [draw, -latex']
\tikzstyle{cloud} = [draw, ellipse,fill=red!20, node distance=3cm,
    minimum height=2em]

\newtheorem{theorem}{Theorem}

\def\sf{{\scriptscriptstyle {f}}}

\def\sRm{{\scriptscriptstyle {R_m}}}
\def\sRm1{{\scriptscriptstyle {R_{m+1}}}}

\def\sCk{{\scriptscriptstyle {C_k}}}

\def\sCk1{{\scriptscriptstyle {C_{k+1}}}}

\def\I3{\mathbf I_{3}}

\newcommand\numberthis{\addtocounter{equation}{1}\tag{\theequation}}

\newcommand*\circled[1]{\tikz[baseline=(char.base)]{
            \node[shape=circle,draw,inner sep=0.5pt] (char) {#1};}}
\newcommand*\colvec[3][]{
    \begin{bmatrix}\ifx\relax#1\relax\else#1\\\fi#2\\#3\end{bmatrix}
}

\begin{document}
\title{Consistent Map-based 3D Localization \\on Mobile Devices
}

\author{Ryan C. DuToit, Joel A. Hesch, Esha D. Nerurkar, and Stergios I. Roumeliotis}

\maketitle

\begin{abstract}

The objective of this paper is to provide \emph{consistent}, real-time 3D localization capabilities to mobile devices navigating within previously mapped areas. 
To this end, we introduce the Cholesky-Schmidt-Kalman filter (C-SKF), which explicitly considers the uncertainty of the prior map, by employing the {\em sparse} Cholesky factor of the map's Hessian, instead of its dense covariance--as is the case for the Schmidt-Kalman filter (SKF).
By doing so, the C-SKF has memory requirements typically linear in the size of the map, as opposed to quadratic for storing the map's covariance.
Moreover, and in order to bound the processing needs of the C-SKF (between linear and quadratic in the size of the map), we introduce a relaxation of the C-SKF algorithm, the sC-SKF, which operates on the Cholesky factors of independent sub-maps resulting from dividing the trajectory and observations used for constructing the map into overlapping segments.
Lastly, we assess the processing and memory requirements of the proposed C-SKF and sC-SKF algorithms, and compare their positioning accuracy against other approximate map-based localization approaches that employ measurement-noise-covariance inflation to compensate for the map's uncertainty.
\end{abstract}

\IEEEpeerreviewmaketitle
\section{Introduction}
In many applications (e.g., surveillance, manufacturing, virtual and augmented reality), robots or people need to accurately localize within a frequently-visited indoor space. 
In such cases, the accuracy and efficiency of localization can be significantly improved by using a map of the area of operation.
In the context of 3D visual-inertial localization, maps computed beforehand\footnote{Besides batch least squares (BLS), pose-graphs~\cite{konolige08tro}, and PTAM~\cite{klein2007parallel} have also been used for reducing the processing cost of map building. Since such approximations yield {\em inconsistent} maps, we do not consider them further in the context of this work.}
have been employed by localization algorithms,%
\footnote{Since we are interested in continuous localization, we do not consider vision-only methods that provide pose estimates only intermittently (e.g.,~\cite{clemens09}).} 
such as the multi-state constraint Kalman filter (MSCKF) in~\cite{Lynen2015} and~\cite{Mourikis09}, and parallel tracking and mapping (PTAM) in~\cite{Middelberg2014} and~\cite{Ventura2014},
to improve positioning accuracy based on visual observations of mapped features.
These methods achieve real-time performance but are not consistent (in this work, we refer to a consistent estimator as having zero-mean error with covariance equal to or larger than the estimation error's true covariance).
Specifically, ~\cite{Middelberg2014} and~\cite{Ventura2014} are inconsistent not only because the assumption of a perfect map but also due to the approximations invoked by the optimization algorithm used for localization; thus they cannot provide a reliable measure of their positioning uncertainty.
On the other hand, \cite{Mourikis09},  which also assumes that the map is perfect and~\cite{Lynen2015}, which ignores the correlations between the estimated state and the map, inflate the camera measurement's noise covariance so as to reduce the effect of inconsistency: overly confident and often unreliable estimates. 
Nevertheless, inflating the measurement noise does not alleviate the inconsistency that arises from ignoring cross-correlations between the estimated state and the map.

An alternative, approximate method, which explicitly accounts for the map's uncertainty and its correlations with the estimated state, is the  Schmidt-Kalman filter (SKF)~\cite{schmidt1966, Simon-textbook}. 
The SKF has processing requirements {\em linear} in the map's size, as it only needs to update the device's state, covariance, and cross-correlation with the map.
Although the map's state and covariance need not be updated, storing its covariance has cost {\em quadratic} in the number of features.
This has been the main drawback of the SKF, as well as of its variants applied to simultaneous localization and mapping (e.g.,~\cite{guivant2001optimization, Julier2001}), which has restricted its use to small-size areas.

To overcome this limitation, in this work, we introduce the Cholesky (C)-SKF which has, typically, {\em linear} in the map's size memory requirements, while providing the same {\em consistency} properties as the SKF.
The key insight behind our approach is that most current methods employed for constructing large-scale maps, such as batch least squares (BLS), compute and use the Cholesky factor of the problem's Hessian, which is sparse [typically linear number of non-zero elements (nnz) in the size of the map\footnote{In extreme cases, such as when the map is constructed using images taken while hovering over the same scene, the memory requirements may become quadratic. In such cases, alternative approaches, such as PTAM, should be employed instead for localization.}] instead of the dense covariance matrix~\cite{Triggs00}. 
Additionally, and in order to reduce the processing requirements  of the C-SKF -- between linear and quadratic in the map's size -- we introduce a {\em consistent} relaxation of the C-SKF, the sub-map (s)C-SKF, which trades localization accuracy for processing speed by operating on the Cholesky factors of the partitioned Hessians resulting from dividing the original map into independent sub-maps. Note that the sub-maps used throughout this work are generated from the method of~\cite{GUO_CM_LINE}, however, other methods that produce sub-maps (i.e.,~\cite{choudhary2015}) could be employed as well.
This approximation allows mapping larger areas and/or operating on resource-constrained mobile devices, such as cell phones and tablets.

In summary, the main contributions of this paper are:

\begin{itemize}
 \item We introduce the Cholesky-Schmidt-Kalman filter (C-SKF), which employs the Hessian's Cholesky factor to compactly represent the map's uncertainty, and efficiently compute {\em consistent} map-based updates.
\item We introduce the sub-map (s)C-SKF, a relaxation of the C-SKF, which employs multiple, independent sub-maps of the area of interest to support real-time, consistent map-based localization on mobile devices.
 \item We validate the accuracy and consistency of the C-SKF and sC-SKF using visual and inertial measurements from mobile devices against VICON ground truth.
\end{itemize}

In what follows, we provide an overview of our map-based localization system (Sect.~\ref{sec:overview}) and then present the system state and measurement models (Sect.~\ref{sec:meas}). 
In Sect.~\ref{sec:alg}, we describe the limitations of the SKF when applied to map-based localization, and then introduce the C-SKF and the sC-SKF.
Our method for generating reliable 2D-3D correspondences is presented in Sect.~\ref{corr_pipeline}.
Lastly, we experimentally validate the proposed algorithms in Sect.~\ref{sec:exp} and provide concluding remarks with a discussion on future work in Sect.~\ref{sec:conclusion}.

\section{Map-based Localization Algorithm Overview} \label{sec:overview}
Our objective is to design a \emph{consistent} estimator for computing, in real-time, the 3D position and attitude (pose) of a mobile device using inertial and visual measurements, as well as a prior map of the area of operation.
To do so, we require the following information from the (offline) mapping process:
\begin{itemize}
\item The Cholesky factor of the (sub)map's Hessian.\footnote{In this work, we employ the cooperative mapping (CM) algorithm of~\cite{GUO_CM_LINE} since, in addition to computing the Cholesky factor of the map's Hessian, it also provides a convenient mechanism for handling sub-maps and computing their corresponding Cholesky factors needed for the sC-SKF.}
\item The 3D position estimates of the mapped features along with their descriptors (e.g., FREAKs~\cite{FREAK} or ORBs~\cite{ORB}) and the vectors of their corresponding images as indexed by a vocabulary tree (VT)~\cite{Nister062}.
\end{itemize}
The former is necessary for representing the map's uncertainty, while the latter is used for recognizing mapped features and using their position estimates for updating the device's pose.

\begin{figure}
\centering
\includegraphics[width=0.47\textwidth]{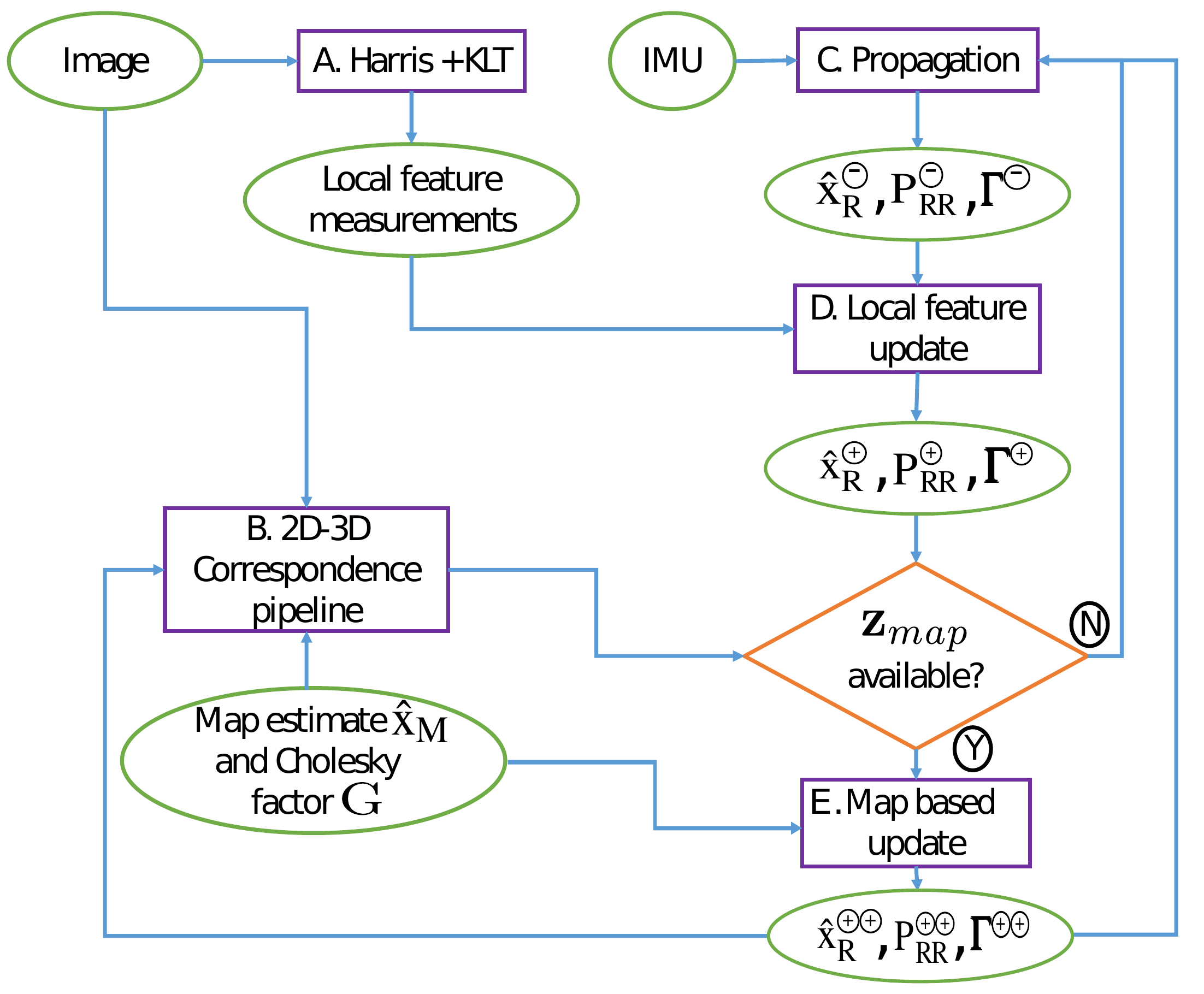}
\caption{Cholesky-Schmidt-Kalman filter (sub)map-based localization algorithm overview.}\label{fig:scheme}
\end{figure}

Given this information, in Fig.~\ref{fig:scheme}, we provide an overview of the (online) image processing and estimation components of our (s)C-SKF map-based localization algorithms.
In particular, at the core of our estimator is the MSCKF~\cite{Guo-RSS-14, Mourikis09} which processes inertial measurements for propagating the device's state and covariance estimates.
Intermittently, and when feature tracks (e.g., Harris corners~\cite{Harris1988} corners tracked by KLT~\cite{Kanade1981}) become available, the proposed (s)C-SKF adaptation of the MSCKF consistently updates only the device's state-covariance and correlations with the map, but {\em not} the map itself (Sect.~\ref{sec:ekf}).
Similarly, every time the 2D-to-3D feature-matching pipeline (Sect.~\ref{corr_pipeline}) finds correspondences between the, e.g., FREAK features extracted in the current image and those found in the map, the (s)C-SKF uses this information to, again, update all estimated quantities except the map and its covariance (Sect.~\ref{sec:csk}~--~\ref{sec:sub-maps}).

\section{System State and Measurement Models} \label{sec:meas}
\subsection{Device State}
At time-step $k$, the estimated state is\footnote{To simplify the subsequent derivations, we assume the camera and IMU are time synchronized and co-located. In practice, we estimate their extrinsic calibration parameters and time offset following the approaches of~\cite{Mirzaei08} and~\cite{Guo-RSS-14}, respectively.}:
\begin{equation}
 \mathbf{x}_k = \begin{bmatrix} \mathbf{x}_{E}^T & \mathbf{x}_{C_{k-1}}^T& \dots& \mathbf{x}_{C_{k-N}}^T & \mathbf{x}_{\tau}^T \end{bmatrix}^T \label{eq:evolve}
\end{equation}
where $\mathbf{x}_{E}$ is the evolving state of the device:
\begin{equation}
 \mathbf{x}_{E} = \begin{bmatrix} ^{I_{k}}\mathbf{q}_G^T & ^G\mathbf{p}_{I_k}^T & \mathbf{b}_{g_k}^T & ^G\mathbf{v}_{I_k}^T & \mathbf{b}_{a_k}^T &  ^G\mathbf{p}_{I_k}^T \end{bmatrix}^T
\end{equation}
$^{I_k}\mathbf{q}_G$ is the quaternion representation of the global, $\{G\}$, frame's orientation in the IMU's current frame, $\{I_k\}$, $\mathbf{b}_{a_k}$ and $\mathbf{b}_{g_k}$ are the accelerometer and gyroscope biases, respectively, and $^G\mathbf{v}_{I_k}$ and $^G\mathbf{p}_{I_k}$ are the velocity and position of $\{I_k\}$ in $\{G\}$.
In~\eqref{eq:evolve}, $\mathbf{x}_{C_{i}} = \begin{bmatrix} ^{I_i}\mathbf{q}_G^T & ^G\mathbf{p}_{I_i}^T & t_{s_i} \end{bmatrix}^T, \; i=k-N,\dots,k-1$ corresponds to previous IMU poses.
Following~\cite{Mourikis09}, we maintain a sliding window of $N$ such poses so as to process measurements to non-mapped (or local) features without incorporating them into the state vector.

Finally, our problem formulation requires estimating the 4 degree of freedom (d.o.f) transformation between the device's global frame, $\{G\}$, and one or more map's frames of reference, $\{M_i\}$:
\begin{align}
  &\mathbf{x}_{\tau} = \begin{bmatrix} \mathbf{x}_{\tau_1}^T & \mathbf{x}_{\tau_2}^T & \dots & \mathbf{x}_{\tau_L}^T \end{bmatrix}^T,
  &\mathbf{x}_{\tau_i} = \begin{bmatrix} ^{M_i}\phi_G & ^G\mathbf{p}^T_{M_i} \end{bmatrix}^T \label{eq:maptransstate}
\end{align}
where $^G\mathbf{p}_{M_i}$ is the position of $\{M_i\}$ in $\{G\}$, and $^{M_i}\phi_G$ is the rotation about gravity between the two frames. Note that since the roll and pitch angles are observable for any vision-aided inertial navigation system (VINS)~\cite{Hesch_TRO_14}, we can choose the $z$-axis to align with gravity in both the global and map coordinate frames.

\subsection{IMU measurement model}
Following~\cite{Mourikis09}, we propagate the state estimate of the device [see~\eqref{eq:evolve}] by integrating the IMU's rotational velocity and linear acceleration measurements, $\mathbf{u}_k$,
\begin{equation}
 \mathbf{x}_{k+1} = \mathbf{g}(\mathbf{x}_k, \mathbf{u}_k) + \mathbf{w}_k \label{eq:prop_eq}
\end{equation}
where $\mathbf{g}$ is a nonlinear function corresponding to the IMU measurement model and $\mathbf{w}_k$ is zero-mean, Gaussian noise of known covariance~\cite{chatfield1997fundamentals}.
\subsection{Local-feature measurement model}
As the device traverses its environment, it observes and tracks (via KLT~\cite{Kanade1981}) point features (Harris corners~\cite{Harris1988}) that have \emph{not} been mapped. These local features are used as in~\cite{Mourikis09} to provide measurement constraints between the $N+1$ IMU-camera poses maintained in the state vector.

The non-linear and linearized local-feature measurement models are
\begin{align}
  &\label{eq:msckf_meas} \mathbf{z} = \mathbf{h}(\mathbf{x}, \mathbf{p}_f) + \mathbf{n} \; ,
 &\mathbf{r} = \mathbf{H}_R \mathbf{\tilde{x}}_R + \mathbf{H}_f \mathbf{\tilde{p}}_f + \mathbf{n}
\end{align}
where $\mathbf{z}$ is the measurement, $\mathbf{h}$ is the perspective-projection camera measurement model, $\mathbf{r}$ is the linearized measurement residual, $\mathbf{x}_R$ ($\mathbf{\tilde{x}}_R$) is the device state (error-state), $\mathbf{p}_f$ ($\mathbf{\tilde{p}}_f$) is the local feature state (error-state), $\mathbf{H}_R$ and $\mathbf{H}_f$ are the measurement Jacobians corresponding to the device  and feature states, respectively, and $\mathbf{n}$ is zero-mean, Gaussian noise with known covariance, $\mathbf{R}$. 

As in~\cite{Mourikis09}, we marginalize the local feature by projecting $\mathbf{r}$ on the left null space of $\mathbf{H}_f$, $\mathbf{U}$:
\begin{align*}
 \mathbf{r}^o &= \mathbf{H}^o_R \mathbf{\tilde{x}}_R + \mathbf{n}^o \numberthis \label{eq:lns_eq}
\end{align*}
where:
\begin{align*}
 &\mathbf{r}^o = \mathbf{U}^T \mathbf{r} \; , &\mathbf{H}^o_R = \mathbf{U}^T \mathbf{H}_R  \; , & &\mathbf{n}^o = \mathbf{U}^T \mathbf{n}
\end{align*}

The new measurement residual, $\mathbf{r}^o$, and Jacobian, $\mathbf{H}^o_R$, are then (Sect.~\ref{sec:msckf_update}) used for updating the device state estimate.
\subsection{Mapped-feature measurement model} \label{sec:MSCKFUpdate}
When a previously-mapped point-feature,~$f$, (expressed with respect to the IMU-camera frame, $\{I_\lambda^{M_i}\}$, that first observed it during mapping) is detected by the device, we can form the following geometric relationship (see Fig.~\ref{fig:mapmeasurement}):\\
\begin{equation}
 ^{I_k}\mathbf{p}_f = ^{I_k}_{G}\mathbf{C} \left( ^G\mathbf{p}_{M_i} - ^G\mathbf{p}_{I_k} + ^{G}_{M_i}\mathbf{C} \left[ ^{M_i}\mathbf{p}_{I^{M_i}_\lambda} + ^{M_i}_{I^{M_i}_\lambda}\mathbf{C} \;\; ^{I^{M_i}_\lambda}\mathbf{p}_{f}\right] \right)
\end{equation}
where all rotation matrices, $\mathbf{C}$, are parameterized by their corresponding 3 d.o.f quaternions, except $ ^{G}_{M_i}\mathbf{C}$, which corresponds to a rotation about gravity by an angle $^{M_i}\phi_G$. Applying the camera perspective-projection transformation, $\bm{\pi}$, leads to the following measurement model:
\begin{figure}
\centering
 \includegraphics[width = 0.5\textwidth]{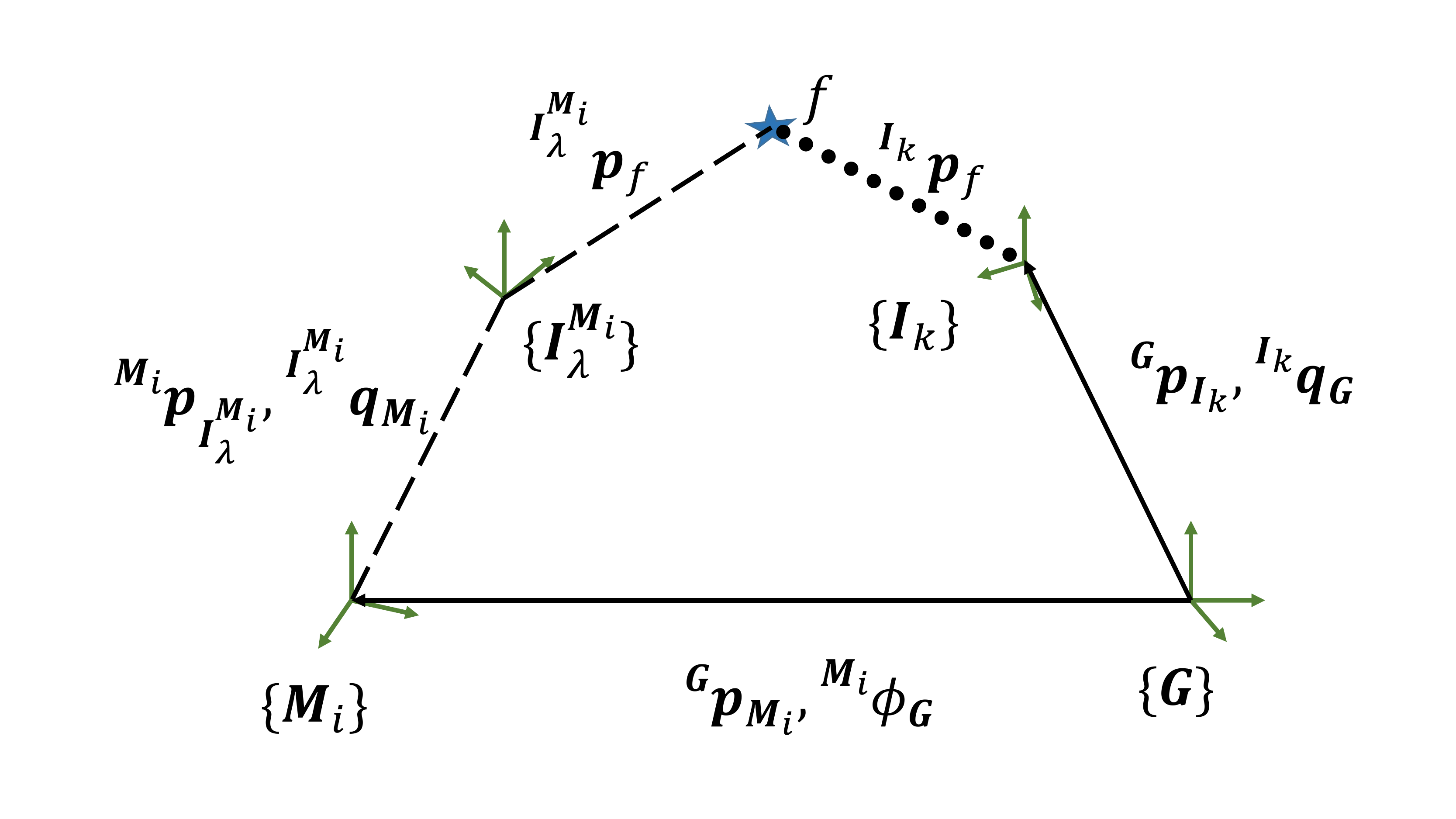}
 \caption{The geometric relationship of a mapped feature observation. The dotted line is a bearing measurement, solid lines correspond to quantities estimated online. Dashed lines correspond to variables determined during mapping offline.}
 \label{fig:mapmeasurement}
\end{figure}
\begin{align*}
 \mathbf{z}_{map} &= \bm{\pi} \left(^{I_k}\mathbf{p}_f\right) + \mathbf{n} \\
  &= \mathbf{h}_{map}(\mathbf{x}_k, ^{I^{M_i}_\lambda}\mathbf{p}_{f}) + \mathbf{n} \numberthis \label{eq:mapmeasurement}
\end{align*}
where $\mathbf{n}$ is zero-mean, Gaussian noise with covariance $\mathbf{R}$.

To simplify the notation, from here on, we omit time indices and denote the state $\mathbf{x}_k$ in~\eqref{eq:evolve} by $\mathbf{x}_R$, while we use $\mathbf{x}_M$ to represent the vector comprising of all mapped features and IMU-camera poses.
Linearizing~\eqref{eq:mapmeasurement} yields:
\begin{align}
 \mathbf{r} = \mathbf{H}_R \mathbf{\tilde{x}}_R + \mathbf{H}_M\mathbf{\tilde{x}}_M + \mathbf{n} \label{eq:mapmeasurementlin}
\end{align}
where $\mathbf{H}_R$ and $\mathbf{H}_M$ are the device and map Jacobians, respectively. Note that both $\mathbf{H}_R$ and $\mathbf{H}_M$ are sparse, as the measurement equation only involves the position and quaternion of the current and mapped IMU-camera pairs, the 4 d.o.f map-to-global transformation, and the position of the feature.

With our system state and measurement models defined, in what follows, we present how measurements \eqref{eq:prop_eq},~\eqref{eq:lns_eq},~and~\eqref{eq:mapmeasurement} are processed in a consistent manner.
\section{Algorithm Description} \label{sec:alg}
In what follows, we first review the SKF, and then introduce the C-SKF and sC-SKF.
To simplify notation, we denote updated values with $^{\circled{\tiny{+}}}$ and propagated values with $^{\circled{\text{-}}}$, rather than using time subscripts.
[i.e., $\mathbf{P}_{k+1|k+1} = \mathbf{f}(\mathbf{P}_{k+1|k})$ is expressed as $\mathbf{P}^{\circled{\tiny{+}}} = \mathbf{f}({\mathbf{P}})$, and $\mathbf{P}_{k+1|k} = \mathbf{g}(\mathbf{P}_{k|k})$ is $\mathbf{P}^{\circled{\text{-}}} = \mathbf{g}({\mathbf{P}})$  ]

\subsection{Background: Schmidt-Kalman filter}\label{sec:skf}
Consider the current propagated system covariance, $\mathbf{P}$, and Jacobian, $\mathbf{H}$, to be:

\begin{align} \label{eq:jac}
\mathbf{P} &= \begin{bmatrix} \mathbf{P}_{RR} & \mathbf{P}_{RM} \\ \mathbf{P}_{MR} & \mathbf{P}_{MM} \end{bmatrix}
&\mathbf{H} = \begin{bmatrix} \mathbf{H}_R & \mathbf{H}_M \end{bmatrix}
\end{align}
The state and covariance update equations for the extended Kalman filter (EKF) are:
\begin{align} 
 \mathbf{\hat{x}}^{\circled{\tiny{+}}}  &= \mathbf{\hat{x}} + \mathbf{K} \mathbf{r}\label{eq:stateupdate} \\ 
 \mathbf{P}^{\circled{\tiny{+}}} &= \left( \mathbf{I} - \mathbf{K}\mathbf{H}\right) \mathbf{P} \left( \mathbf{I} - \mathbf{H}^T\mathbf{K}^T\right) + \mathbf{K} \mathbf{R} \mathbf{K}^T \label{eq:covupdate}
 \end{align}
 where
\begin{align*}
 \mathbf{r} &= \mathbf{z} - \mathbf{h}(\mathbf{\hat x}) \\
 \mathbf{S} &= \mathbf{HPH}^T + \mathbf{R} \\
 \mathbf{K} &= \mathbf{P}\mathbf{H}^T\mathbf{S}^{-1} =   \begin{bmatrix}
    \mathbf{P}_{RR}\mathbf{H}_R^T + \mathbf{P}_{RM}\mathbf{H}_M^T \\ \mathbf{P}_{MR}\mathbf{H}_R^T + \mathbf{P}_{MM}\mathbf{H}_M^T
  \end{bmatrix}\mathbf{S}^{-1} \\
  &= \begin{bmatrix}
    \mathbf{\bar{K}}_R  \\ \mathbf{\bar{K}}_M
  \end{bmatrix}\mathbf{S}^{-1}
  = \mathbf{\bar{K}} \mathbf{S}^{-1}\label{eq:kbar} \numberthis
\end{align*}
As shown in~\cite{Simon-textbook}, the SKF updates only part of state by zeroing the Kalman gain associated with the state elements whose estimates are to remain the same (in our case the map, $\mathbf{x}_M$):
\begin{equation} \label{eq:kalmangain}
 \mathbf{K}_{SKF} = \begin{bmatrix} (\mathbf{\bar{K}}_R \mathbf{S}^{-1})^T & \mathbf{0} \end{bmatrix}^T
\end{equation}
Substituting~\eqref{eq:kalmangain} in~\eqref{eq:stateupdate} yields the following state update:
\begin{align}
  &\mathbf{\hat{x}}_R^{\circled{\tiny{+}}}  = \mathbf{\hat{x}}_R + \mathbf{\bar{K}}_R \mathbf{S}^{-1}\mathbf{r} 
  &\mathbf{\hat{x}}_M^{\circled{\tiny{+}}}  = \mathbf{\hat{x}}_M \label{eq:schmidt_state_update}
\end{align}
Additionally, by employing~\eqref{eq:kalmangain} in~\eqref{eq:covupdate}, we arrive at the SKF's covariance update:
\begin{equation} \label{eq:final_schmidt_p}
 \mathbf{P}^{\circled{\tiny{+}}}_{SKF} = \mathbf{P} - \begin{bmatrix} \mathbf{\bar{K}}_R \mathbf{S}^{-1} \mathbf{\bar{K}}_R^T & \mathbf{\bar{K}}_R \mathbf{S}^{-1} \mathbf{\bar{K}}_M^T \\ \mathbf{\bar{K}}_M \mathbf{S}^{-1} \mathbf{\bar{K}}_R^T & \mathbf{0}\end{bmatrix}
\end{equation}

Note that if we express the updated covariance of the EKF as a function of the SKF updated covariance, it is straightforward to show:
\begin{align}
   \mathbf{P}^{\circled{\tiny{+}}}_{EKF} &= \mathbf{P}^{\circled{\tiny{+}}}_{SKF} +  \begin{bmatrix} \mathbf{0} \\ \mathbf{\bar{K}}_M \end{bmatrix} \mathbf{S}^{-1}  \begin{bmatrix} \mathbf{0} & \mathbf{\bar{K}}^T_M \end{bmatrix} \\
   \Rightarrow     \mathbf{P}^{\circled{\tiny{+}}}_{EKF} &\preceq \mathbf{P}^{\circled{\tiny{+}}}_{SKF}
\end{align}
Since $\mathbf{S}^{-1}$ is positive definite, and hence $\begin{bmatrix} \mathbf{0} \\ \mathbf{\bar{K}}_M \end{bmatrix} \mathbf{S}^{-1}  \begin{bmatrix} \mathbf{0} & \mathbf{\bar{K}}^T_M \end{bmatrix} $ must be positive semi-definite.

Thus, we have shown that the SKF is consistent as it does not underestimate the system covariance. Furthermore, as evident from~\eqref{eq:final_schmidt_p}, the cost of an SKF covariance update is linear in the size of the map, as only the device's covariance and cross-correlation terms need to be updated, while $\mathbf{H}$ in~\eqref{eq:jac} is sparse.

On the other hand the SKF requires storing the covariance of the map, $\mathbf{P}_{MM}$, which is dense. To better appreciate the challenge this poses on mobile devices, we note that the covariance of a map generated from 1.5 min of visual and inertial data requires 1.2 GB of storage space. Instead, the sparse Cholesky factor of the corresponding Hessian matrix requires only 76 MB. This motivates us to introduce the Cholesky-SKF (C-SKF) in the next sections.
\subsection{C-SKF device-map initialization} \label{sec:init}
In what follows, we start by describing the process for estimating the 4 d.o.f transformation between the map's frame and the device's global reference frame, as well as its covariance and correlations with all estimated quantities.

Specifically, consider the first time the mobile device observes two or more previously-mapped features. An  initial estimate, $\mathbf{\hat{x}}_{\tau}$, for the uniform 4 d.o.f transformation is obtained by employing the 2+1 pt RANSAC~\cite{kukelova2011}. Furthermore, we partition the current state as
\begin{equation}
\mathbf{x} = \begin{bmatrix} \mathbf{x}_{R'}^T & \mathbf{x}_{\tau}^T & \mathbf{x}_M^T \end{bmatrix}^T
\end{equation}
where $\mathbf{x}_{R'}$ comprises of the remaining elements of the device's state vector [see~\eqref{eq:evolve}], and the corresponding covariance as
\begin{equation}
 \mathbf{P} = \begin{bmatrix} \mathbf{P}_{R'R'} & \mathbf{0} & \mathbf{0} \\
    \mathbf{0} & \mathbf{P}_{\tau\tau} & \mathbf{0} \\
    \mathbf{0} & \mathbf{0} & (\mathbf{G}\mathbf{G}^T)^{-1} \\
  \end{bmatrix}, \mathbf{P}_{\tau\tau} = \lim_{\mu \rightarrow \infty}{(\mu\mathbf{I}_4)} 
\end{equation}
where $\mathbf{P}_{\tau\tau}$ is the covariance of the unknown 4 d.o.f transformation, $\mathbf{GG}^T$ is the Hessian of the map, and $\mathbf{G}$ is its Cholesky factor. After linearizing the measurement model in~\eqref{eq:mapmeasurement} and denoting the corresponding Jacobian as
\begin{equation} 
\mathbf{H} = [\mathbf{H}_{R'} \quad \mathbf{H}_\tau \quad \mathbf{H}_M] \\
\end {equation}
we employ \eqref{eq:kbar}-\eqref{eq:schmidt_state_update} to update the estimates of $\mathbf{x}_{R'}$ and $\mathbf{x}_{\tau}$:
\begin{align}
 \mathbf{\hat{x}}_{R'}^{\circled{\tiny{+}}} &= \mathbf{\hat{x}}_{R'} + \mathbf{P}_{R'R'} \mathbf{H}_{R'}^T \mathbf{S}^{-1} \mathbf{r} \\
  \mathbf{\hat{x}}_{\tau}^{\circled{\tiny{+}}} &= \mathbf{\hat{x}}_{\tau} + (\mathbf{H}_\tau^T\mathbf{A}^{-1}\mathbf{H}_\tau)^{-1}\mathbf{H}_\tau^T\mathbf{A}^{-1}\mathbf{r} \\
  \mathbf{\hat{x}}_{M}^{\circled{\tiny{+}}} &=  \mathbf{\hat{x}}_{M} 
\end{align}
Additionally, employing~\eqref{eq:final_schmidt_p}, it can be shown that the updated SKF covariance is
\begin{equation}
 \mathbf{P} = \begin{bmatrix} \mathbf{P}_{R'R'}^{\circled{\tiny{+}}} & \mathbf{P}_{R'\tau}^{\circled{\tiny{+}}} & \mathbf{P}_{R'M}^{\circled{\tiny{+}}}     \vspace{1mm}
 \\
    \vspace{1mm}
    \mathbf{P}_{R'\tau}^{\circled{\tiny{+}}T}  & \mathbf{P}_{\tau\tau}^{\circled{\tiny{+}}} & \mathbf{P}_{\tau M}^{\circled{\tiny{+}}} \\
    \mathbf{P}_{R'M}^{\circled{\tiny{+}}T} & \mathbf{P}_{\tau M}^{\circled{\tiny{+}}T}  & (\mathbf{G}\mathbf{G}^T)^{-1} \\
  \end{bmatrix} = \begin{bmatrix}  \mathbf{P}_{RR}^{\circled{\tiny{+}}} & \mathbf{P}_{RM}^{\circled{\tiny{+}}} \\ \mathbf{P}_{MR}^{\circled{\tiny{+}}} & (\mathbf{G}\mathbf{G}^T)^{-1}\end{bmatrix} \label{eq:inicov}
\end{equation}
where
\begin{align*}
 \mathbf{P}_{R'R'}^{\circled{\tiny{+}}} &= \mathbf{P}_{R'R'} - \mathbf{P}_{R'R'}\mathbf{H}^T_{R'}\mathbf{S}^{-1}\mathbf{H}_{R'}\mathbf{P}_{R'R'} \\
 \mathbf{P}_{R'\tau}^{\circled{\tiny{+}}} &= - \mathbf{P}_{R'R'}   \mathbf{H}^T_{R'}\mathbf{A}^{-1}\mathbf{H}_\tau (\mathbf{H}_\tau^T\mathbf{A}^{-1}\mathbf{H}_\tau)^{-1} \\
 \mathbf{P}_{\tau \tau}^{\circled{\tiny{+}}} &= (\mathbf{H}_\tau^T \mathbf{A} \mathbf{H}_\tau)^{-1} \\
 \mathbf{P}_{RM}^{\circled{\tiny{+}}} &= \begin{bmatrix} \mathbf{P}_{R'M }^{\circled{\tiny{+}}} \\  \mathbf{P}_{\tau M}^{\circled{\tiny{+}}} \end{bmatrix} 
 =\begin{bmatrix}  \mathbf{P}_{R'R'} - \mathbf{P}_{R'R'}\mathbf{H}^T_{R'}\mathbf{S}^{-1}\mathbf{J} \\  (\mathbf{H}_\tau^T\mathbf{A}^{-1}\mathbf{H}_\tau)^{-1}\mathbf{H}_\tau^T\mathbf{A}^{-1}\mathbf{J} \end{bmatrix} \mathbf{G}^{-1} \\
 &= \bm{\Gamma} \mathbf{G}^{-1} \numberthis \label{eq:gammadef}
 \end{align*}
 and
 \begin{align*}
  \mathbf{S}^{-1} &= \mathbf{A}^{-1} - \mathbf{A}^{-1}\mathbf{H}_\tau(\mathbf{H}_\tau^T\mathbf{A}^{-1}\mathbf{H}_\tau)\mathbf{H}_\tau^T\mathbf{A}^{-1} \\
 \mathbf{A} &= \mathbf{H}_{R'}\mathbf{P}_{R'R'}\mathbf{H}_{R'} + \mathbf{J}\mathbf{J}^T + \mathbf{R}
\end{align*}
Finally, $\mathbf{J}$ is defined as:
\begin{equation}
 \mathbf{G}\mathbf{J}^T = \mathbf{H}_M^T \label{eq:J_def}
\end{equation}

A key element of our approach is that, from this point on, instead of updating the cross-correlation term, $\mathbf{P}_{RM}$, we will represent it in a factorized form [see~\eqref{eq:gammadef}] and apply updates on its factor, $\bm{\Gamma}$, as is shown in Sect.~\ref{sec:ekf} and Sect.~\ref{sec:csk}. Following this convention, we have:
\begin{equation}
\mathbf{P} = 
\begin{bmatrix} 
    \mathbf{P}_{RR} & \bm{\Gamma} \mathbf{G}^{-1} \\
    (\bm{\Gamma}\mathbf{G}^{-1})^T  & (\mathbf{G}\mathbf{G}^T)^{-1} \label{eq:cov_similar_to_msckf}
\end{bmatrix}
\end{equation}

Note also that we do not need to compute the inverse of the Cholesky factor, $\mathbf{G}$. Instead, all update equations involving $\mathbf{G}$ will be of the same form as~\eqref{eq:J_def}, where a back-solve involving the sparse $\mathbf{G}$ is required for efficiently computing $\mathbf{J}^T$.

\subsection{C-SKF propagation and local-feature update}\label{sec:ekf}

\subsubsection{IMU-based Propagation}
By employing the IMU measurement model in~\eqref{eq:prop_eq}, the device's state is propagated while, as is the case for the SKF, the map's state remains the same:
\begin{align}
 &\mathbf{\hat{x}}^{\circled{\text{-}}}_R = \mathbf{f}(\mathbf{\hat{x}}_R, \mathbf{u}) \;,
 &\mathbf{\hat{x}}^{\circled{\text{-}}}_M = \mathbf{x}_M 
\end{align}
On the other hand, and in order to propagate the covariance of the C-SKF [see~\eqref{eq:cov_similar_to_msckf}], we employ the EKF covariance propagation equation:
\begin{align}
\mathbf{P}^{\circled{\text{-}}} &= \bm{\Phi}_{C} \mathbf{P} \bm{\Phi}_{C}^T + \mathbf{Q}_{C}
 = \begin{bmatrix} \bm{\Phi}\mathbf{P}_{RR}\bm{\Phi}^T + \mathbf{Q} & \bm{\Phi}\bm{\Gamma}\mathbf{G}^{-1} \\ (\bm{\Phi}\bm{\Gamma}\mathbf{G}^{-1})^T & (\mathbf{GG}^T)^{-1} \end{bmatrix} \numberthis \label{eq:cskfprop}
\end{align}
with
\begin{align}
 &\bm{\Phi}_{C} = \begin{bmatrix} \bm{\Phi} &  \mathbf{0} \\ \mathbf{0} & \mathbf{I} \end{bmatrix}\;,
 &\mathbf{Q}_{C} = \begin{bmatrix} \mathbf{Q} &  \mathbf{0} \\ \mathbf{0} & \mathbf{0} \end{bmatrix}
\end{align}
where $\bm{\Phi}$ and $\mathbf{Q}$ are the IMU Jacobian of the state and corresponding noise covariance, respectively.

As evident from~\eqref{eq:cskfprop}, the device's propagated covariance is computed at low cost, while the cross-correlation terms only require modifying the cross-correlation factor, $\bm{\Gamma}$, (i.e., $\bm{\Gamma}^{\circled{\text{-}}}  = \bm{\Phi}\bm{\Gamma}$) at cost linear in the size of the map.
\subsubsection{C-SKF Local-Feature Measurement Update}\label{sec:msckf_update}
When a local-feature-track measurement becomes available [see~\eqref{eq:lns_eq}], we arrive at the following Jacobian:
\begin{equation}
 \mathbf{H} = \begin{bmatrix} \mathbf{H}^o_R & \mathbf{0} \end{bmatrix}
\end{equation}
As in~\eqref{eq:kalmangain}, we zero out the Kalman gain associated with the mapped states, leading to an update of the device state, while the map remains the same:
\begin{align}
 &\mathbf{\hat{x}}_R^{\circled{\tiny{+}}} = \mathbf{\hat{x}}_R + \mathbf{P}_{RR} \mathbf{H}^{oT}_R\mathbf{S}^{-1} \mathbf{r} \;,
 &\mathbf{\hat{x}}_M^{\circled{\tiny{+}}} = \mathbf{\hat{x}}_M \label{eq:msckf_state_update}
\end{align}
For the covariance update, we first denote $\mathbf{\bar{K}} = \mathbf{P}\mathbf{H}$:
\begin{equation}
\begin{bmatrix} \mathbf{\bar{K}}_R \\ \mathbf{\bar{K}}_M \end{bmatrix}
  = \begin{bmatrix}
      \mathbf{P}_{RR} & \bm{\Gamma}\mathbf{G}^{-1} \\
      (\bm{\Gamma}\mathbf{G}^{-1})^T & (\mathbf{GG}^T)^{-1}
    \end{bmatrix}
    \begin{bmatrix}
      \mathbf{H}_R^{oT} \\ \mathbf{0}
    \end{bmatrix}
  = \begin{bmatrix}
      \mathbf{P}_{RR}\mathbf{H}_R^{oT} \\
      \mathbf{G}^{-T}\bm{\Gamma}^T\mathbf{H}_R^{oT}
    \end{bmatrix} \label{eq:newkbar}
\end{equation}
Next, we employ~\eqref{eq:newkbar} and~\eqref{eq:cov_similar_to_msckf} in~\eqref{eq:final_schmidt_p} without evaluating $\mathbf{\bar{K}}_M$, yielding:
\begin{align}
&\mathbf{P}_{RR}^{\circled{\tiny{+}}}  = \mathbf{P}_{RR} - \mathbf{P}_{RR}\mathbf{H}_R^{oT}\mathbf{S}^{-1}\mathbf{H}^o_R\mathbf{P}_{RR} \;,
&\mathbf{P}_{MM}^{\circled{\tiny{+}}}  &= \mathbf{P}_{MM}
\end{align}
and
\begin{align*}
\mathbf{P}_{RM}^{\circled{\tiny{+}}} &=  \bm{\Gamma}\mathbf{G}^{-1} - \mathbf{\bar{K}}_R\mathbf{S}^{-1}\mathbf{\bar{K}}_M^T \\
 &=\left(\mathbf{I} - \mathbf{P}_{RR}\mathbf{H}_R^{oT}\mathbf{S}^{-1}\mathbf{H}^o_R\right) \bm{\Gamma}\mathbf{G}^{-1} \\
 &= \bm{\Gamma}^{\circled{\tiny{+}}}\mathbf{G}^{-1} \numberthis \label{eq:FactorizedFormCovariance}
\end{align*}

As evident from~\eqref{eq:msckf_state_update}--\eqref{eq:FactorizedFormCovariance}, the local-feature update has complexity \emph{linear} in the size of the map, as we only need to apply a standard EKF update on the device's covariance, and update the cross-correlation factor, $\bm{\Gamma}$.
\subsection{C-SKF map-based updates}\label{sec:csk}
When the device observes previously mapped features (Sect.~\ref{corr_pipeline}), we employ the methodology of Sect.~\ref{sec:skf}~[\eqref{eq:kbar}--\eqref{eq:final_schmidt_p}] using the measurement model of~\eqref{eq:mapmeasurement}--\eqref{eq:mapmeasurementlin}, and operate on the system covariance with factorized cross-correlation, as defined in~\eqref{eq:cov_similar_to_msckf}.
Specifically, we denote $\mathbf{\bar{K}} = \mathbf{P}\mathbf{H}^T$ as:
  \begin{equation} \label{eq:kbar2}
    \begin{bmatrix} 
      \mathbf{\bar{K}}_R  \\ \mathbf{\bar{K}}_M
    \end{bmatrix}    = \begin{bmatrix} 
      \mathbf{P}_{RR}\mathbf{H}_R^T + \bm{\Gamma}\mathbf{G}^{-1}\mathbf{H}_M^T \\ \mathbf{G}^{-T}\bm{\Gamma}^T\mathbf{H}_R^T + (\mathbf{G}\mathbf{G}^T)^{-1}\mathbf{H}_M^T 
    \end{bmatrix} 
  \end{equation}
Note that we do not explicitly compute $\mathbf{\bar{K}}_M$, instead we first compute $\mathbf{J}$ as:
\begin{equation} \label{eq:Jdef}
\mathbf{G} \mathbf{J}^T  = \mathbf{H}_M^{T}
\end{equation}
Because $\mathbf{G}$ is triangular, we can compute $\mathbf{J}$ with a back-solve operation.
Substituting $\mathbf{J}$ into (\ref{eq:kbar2}) yields:
  \begin{equation} \label{eq:kbarsub}
    \begin{bmatrix} 
      \mathbf{\bar{K}}_R  \\ \mathbf{\bar{K}}_M
    \end{bmatrix} 
    = \begin{bmatrix} 
      \mathbf{P}_{RR}\mathbf{H}_R^T + \bm{\Gamma}\mathbf{J}^T \\ \mathbf{G}^{-T}\bm{\Gamma}^T\mathbf{H}_R^T + \mathbf{G}^{-T}\mathbf{J}^T
    \end{bmatrix}
  \end{equation}
Next, we compute the residual covariance:
\begin{equation}
  \mathbf{S} = \mathbf{H}_R\mathbf{P}_{RR}\mathbf{H}_R^T
  + \mathbf{H}_R\bm{\Gamma}\mathbf{J}^T
  + \mathbf{J}\bm{\Gamma}^T\mathbf{H}_R^T
  + \mathbf{J}\mathbf{J}^T
  + \mathbf{R}
\end{equation}
and state update [see~\eqref{eq:schmidt_state_update}]:
\begin{align*}
  &\mathbf{\hat{x}}_R^{\circled{\tiny{+}}}  = \mathbf{\hat{x}}_R + \mathbf{\bar{K}}_R\mathbf{S}^{-1}\mathbf{r} \; ,
  &\mathbf{\hat{x}}_M^{\circled{\tiny{+}}} = \mathbf{\hat{x}}_M \numberthis
\end{align*}
Finally, we update the covariance using~\eqref{eq:final_schmidt_p} with~\eqref{eq:kbarsub}. Avoiding the calculation of $\mathbf{\bar{K}}_M$, and following the same process in \eqref{eq:FactorizedFormCovariance}, we produce a factorized form of the updated cross-correlation:
\begin{align*}
  \mathbf{P}_{RR}^{\circled{\tiny{+}}} 
  &= \mathbf{P}_{RR}
  - \mathbf{\bar{K}}_R\mathbf{S}^{-1}\mathbf{\bar{K}}_R^T \\
\mathbf{P}_{RM}^{\circled{\tiny{+}}} 
  &= \bm{\Gamma}\mathbf{G}^{-1}
  - \mathbf{\bar{K}}_R\mathbf{S}^{-1}
    (\mathbf{H}_R\bm{\Gamma}\mathbf{G}^{-1}
      + \mathbf{J}\mathbf{G}^{-1}) \\
  &= [\bm{\Gamma}
  - \mathbf{\bar{K}}_R\mathbf{S}^{-1}
    (\mathbf{H}_R\bm{\Gamma}
      + \mathbf{J})]\mathbf{G}^{-1} \\
  &= \bm{\Gamma}^{\circled{\tiny{+}}}\mathbf{G}^{-1} \numberthis \label{eq:crosscorrelation}
\end{align*}

At this point, we should note that unlike propagation and local-feature updates, the processing requirements of mapped-feature updates are not strictly linear in the size of the map. The main bottleneck is~\eqref{eq:Jdef}, as computing $\mathbf{J}$ has complexity between linear and quadratic in the size of the map (depending on the structure of $\mathbf{G}$). As shown in Sect.~\ref{sec:exp}, as the map grows, the time to compute $\mathbf{J}$ becomes unacceptable for real-time operation on a mobile device. This limitation motivates us to introduce the sub-map relaxation of Sect.~\ref{sec:sub-maps-def} and Sect.~\ref{sec:sub-maps}. That is, we decrease the size of $\mathbf{G}$ in~\eqref{eq:Jdef} by partitioning the map into independent sub-maps, while retaining consistency.

\subsection{Sub-map relaxation} \label{sec:sub-maps-def}
\begin{figure}[!ht]
\centering
 \vspace{-10mm}
 \includegraphics[width=0.55\textwidth]{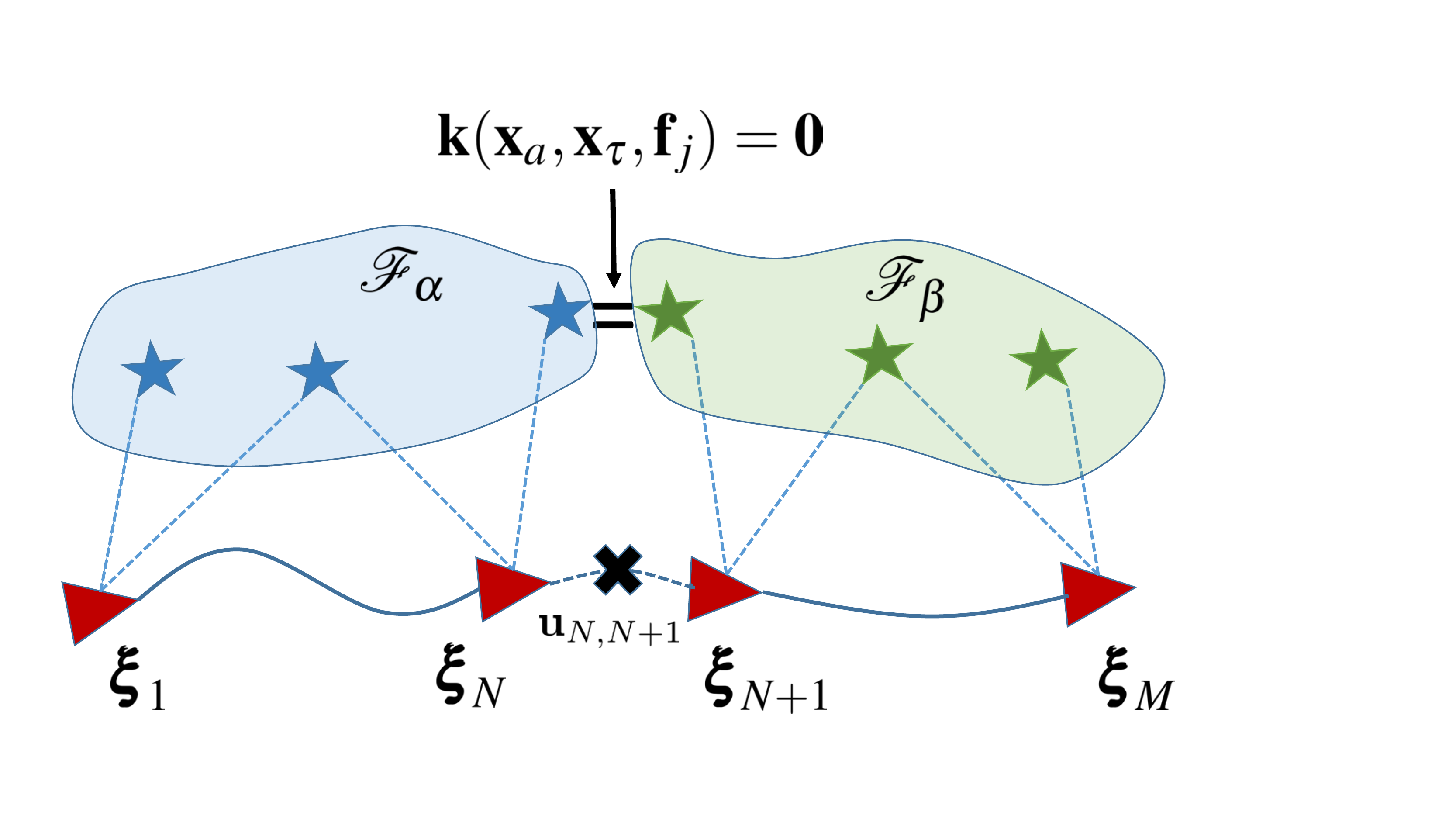}
 \caption{A partitioning of a single map into two sub-maps, with a geometric constraint, $\mathbf{k}(\mathbf{x}_a, \mathbf{x}_\tau, \mathbf{f}_j) = \mathbf{0}$, imposed to common features.} \label{fig:submap}
\end{figure}

Before discussing the sC-SK, we first describe the sub-mapping relaxation process. As shown in Fig.~\ref{fig:submap} and described below, we divide the trajectory and associated features into two separate sets:\footnote{To simplify the explanation, without loss of generality, we describe the case for two sub-maps}

First, the IMU-camera poses $\bm{\xi}_i$, are divided into two sets ([$\bm{\xi}_1$ to $\bm{\xi}_N$] and [$\bm{\xi}_{N+1}$ to $\bm{\xi}_M$]). Our current implementation evenly distributes these sets in time. As part of our future work, we seek to find an optimal partitioning into sub-maps. With this division defined, we partition features into two sets: (i)~$\mathcal{F}_\alpha$: those observed by IMU-camera poses $\bm{\xi}_1$ to $\bm{\xi}_N$, and (ii)~$\mathcal{F}_\beta$: those observed by IMU-camera poses $\bm{\xi}_{N+1}$ to $\bm{\xi}_M$. Features in $\mathcal{F}_\alpha \cap  \mathcal{F}_\beta$ are common features appearing in both sub-maps. The cooperative mapping (CM) algorithm~\cite{GUO_CM_LINE} ``duplicates'' these features such that the two sub-map's individual cost functions are independent, but also introduces a non-linear constraint between these common features
\begin{equation}
\mathbf{k}(\mathbf{x}_a, \mathbf{x}_\tau, \mathbf{f}_j) = \mathbf{0}, \; \mathbf{f}_j \in  \mathcal{F}_\alpha \cap
\mathcal{F}_\beta
\end{equation} where $\mathbf{x}_a$ is the state of all IMU-camera poses and $\mathbf{x}_\tau$ is the 4 d.o.f transformation between the two sub-maps. For each common feature, this constraint enforces its corresponding ``duplicate'' in $\mathcal{F}_\alpha$ or $\mathcal{F}_\beta$ to have the same physical position. Finally, all camera and IMU measurements, $\mathbf{z}_{i,j}$ and $\mathbf{u}_{i,i+1}$, are assigned to their corresponding sub-map (with the execption of $\mathbf{u}_{N,N+1}$, which is discarded).

With such a partitioning, by ignoring the common-feature constraints, we can form two independent cost functions corresponding to each sub-map, $\mathcal{C}_1$ and $\mathcal{C}_2$:
\begin{align*}
 \mathcal{C}_1 &= \sum_{\substack{i=1 \\ \mathbf{f}_j \in \mathcal{F}_\alpha} }^N {||\mathbf{z}_{i,j} - \mathbf{h}(\bm{\xi}_i, \mathbf{f}_j)||_\mathbf{R}} + \sum_{i=1}^{N-1}{||\bm{\xi}_{i+1} - \mathbf{g}(\bm{\xi}_i, \mathbf{u}_{i,i+1})||_\mathbf{Q}}  \\ \numberthis \label{eq:sub_cost}
 \mathcal{C}_2 &= \sum_{\substack{i=N+1 \\ \mathbf{f}_j \in \mathcal{F}_\beta} }^M {||\mathbf{z}_{i,j} - \mathbf{h}(\bm{\xi}_i, \mathbf{f}_j)||_\mathbf{R}} + \sum_{i=N+1}^{M-1}{||\bm{\xi}_{i+1} - \mathbf{g}(\bm{\xi}_i, \mathbf{u}_{i,i+1})||_\mathbf{Q}}
\end{align*}
where $\mathbf{g}$ and $\mathbf{h}$ are defined in~\eqref{eq:msckf_meas} and~\eqref{eq:prop_eq}, respectively.
Summing these two cost functions and imposing the common feature constraints yields:
\begin{align*} \label{eq:cm_cost}
 \mathcal{C} &= \mathcal{C}_1 + \mathcal{C}_2 \numberthis \\ 
  \text{s. t.} & \quad \mathbf{k}(\mathbf{x}_a, \mathbf{x}_\tau, \mathbf{f}_j) = \mathbf{0}, \; \mathbf{f}_j \in  \mathcal{F}_\alpha \cap  \mathcal{F}_\beta
\end{align*}               
We minimize~\eqref{eq:cm_cost} by employing the CM algorithm of~\cite{GUO_CM_LINE}.

\subsection{Cholesky-Kalman-Schmidt with sub-maps (sC-SKF)} \label{sec:sub-maps}
In our problem, we take advantage of the high-accuracy estimate computed from the solution of~\eqref{eq:cm_cost}, but relax the information attributed to each sub-map by storing the Cholesky factors, $\mathbf{G}_i$, resulting from each corresponding cost function in~\eqref{eq:sub_cost} linearized at the CM solution of~\eqref{eq:cm_cost}. Such a relaxation causes the sub-maps to become independent, but maintains consistency.
\begin{theorem} \label{th}
The covariance of each sub-map when ignoring the common-feature constraints is larger or equal, in the positive semi-definite sense, to that computed from the CM.
\end{theorem}
\begin{IEEEproof}
 See Appendix~\ref{cm_appen} for proof of Theorem~\ref{th}.
\end{IEEEproof}

To process mapped measurements with sub-maps, we represent the system covariance as:
\begin{align}
\mathbf{P} 
  = \begin{bmatrix} 
    \mathbf{P}_{RR} & \bm{\Gamma}_1\mathbf{G}_1^{-1}& \bm{\Gamma}_2\mathbf{G}_2^{-1}
    \\ (\bm{\Gamma}_1\mathbf{G}_1^{-1})^T &(\mathbf{G}_1\mathbf{G}_1^T)^{-1} & \mathbf{0} 
    \\  (\bm{\Gamma}_2\mathbf{G}_2^{-1})^T & \mathbf{0} & (\mathbf{G}_2\mathbf{G}_2^T)^{-1}
  \end{bmatrix}
  \end{align}
  where $\bm{\Gamma}_i$ and $\mathbf{G}_i$ is the cross-correlation factor from~\eqref{eq:cov_similar_to_msckf} and the Cholesky factor of the $i$'th sub-map, respectively.

When a mapped feature in the first sub-map is observed (without loss of generality), the measurement Jacobian is:
\begin{align}
  \mathbf{H} = \begin{bmatrix} \mathbf{H}_R & \mathbf{H}_1  & \mathbf{0} \end{bmatrix}
\end{align}
where $\mathbf{H}_1$ is the measurement Jacobian corresponding to the states in the first sub-map.
We compute $\mathbf{J}_1$ and denote $\mathbf{\bar{K}}$ in similar fashion as in~\eqref{eq:Jdef} and~\eqref{eq:kbarsub}, respectively:

  \begin{align} \label{eq:Jsub}
   &\mathbf{G}_1 \mathbf{J}_1^T = \mathbf{H}^T_1 \;, 
   & \begin{bmatrix} 
      \mathbf{\bar{K}}_R  
      \\ \mathbf{\bar{K}}_1
      \\ \mathbf{\bar{K}}_2
    \end{bmatrix} 
    &= \begin{bmatrix} 
      \mathbf{P}_{RR}\mathbf{H}_R^T + \bm{\Gamma}_1\mathbf{J}_1^T 
      \\ \mathbf{G}_1^{-T}(\bm{\Gamma}_1^T\mathbf{H}_R^T + \mathbf{J}_1^T) 
      \\  \mathbf{G}_2^{-T}\bm{\Gamma}_2^T\mathbf{H}_R^T
    \end{bmatrix} 
  \end{align}
The residual covariance and state update are computed as:
\begin{equation*}
  \mathbf{S} = \mathbf{H}_R\mathbf{P}_{RR}\mathbf{H}_R^T
  + \mathbf{H}_R\bm{\Gamma}_1\mathbf{J}_1^T
  + \mathbf{J}_1\bm{\Gamma}_1^T\mathbf{H}_R^T
  + \mathbf{J}_1\mathbf{J}_1^T
  + \mathbf{R}
\end{equation*}
\begin{align}
  \mathbf{\hat{x}}^{\circled{\tiny{+}}}_R &= \mathbf{\hat{x}}_R + \mathbf{\bar{K}}_R\mathbf{S}^{-1}\mathbf{r} \; ,
  &\mathbf{\hat{x}}^{\circled{\tiny{+}}}_1 = \mathbf{\hat{x}}_1 \;, &&\mathbf{\hat{x}}^{\circled{\tiny{+}}}_2 = \mathbf{\hat{x}}_2
\end{align}
Finally, we update the device covariance and cross-correlation using the same factorization as the single-map case [see~\eqref{eq:crosscorrelation}]:
\begin{align*}
  \mathbf{P}_{RR}^{\circled{\tiny{+}}} 
  &= \mathbf{P}_{RR}
  - \mathbf{\bar{K}}_R\mathbf{S}^{-1}\mathbf{\bar{K}}_R^T \;,
 &\mathbf{P}_{11}^{\circled{\tiny{+}}}
  = \mathbf{P}_{11} \;, 
   & &\mathbf{P}_{22}^{\circled{\tiny{+}}}    = \mathbf{P}_{22}
   \end{align*}
   \begin{align*}
  \mathbf{P}_{R1}^{\circled{\tiny{+}}}
  &= \bm{\Gamma}_1\mathbf{G}_1^{-1}
  - \mathbf{\bar{K}}_R\mathbf{S}^{-1}\mathbf{\bar{K}}_1^T \\
  &= [\bm{\Gamma}_1
  - \mathbf{\bar{K}}_R\mathbf{S}^{-1}
    (\mathbf{H}_R\bm{\Gamma}_1
      + \mathbf{J}_1)]\mathbf{G}_1^{-1} \\
  &= \bm{\Gamma}_1^{\circled{\tiny{+}}}\mathbf{G}_1^{-1} \numberthis \\
  \mathbf{P}_{R2}^{\circled{\tiny{+}}} &=   \bm{\Gamma}_2\mathbf{G}_2^{-1}
  - \mathbf{\bar{K}}_R\mathbf{S}^{-1}\mathbf{\bar{K}}_2^T \\
  &= [\mathbf{I} - \mathbf{\bar{K}}_R\mathbf{S}^{-1}
    \mathbf{H}_R] \bm{\Gamma}_2\mathbf{G}_2^{-1} \\
  &=  \bm{\Gamma}_2^{\circled{\tiny{+}}}\mathbf{G}_2^{-1} \numberthis
\end{align*}

By employing the sub-map relaxation, we have significantly lowered the computational requirements of the C-SKF. Note again that the bottleneck of our system is the computation of $\mathbf{J}_1$ [see~\eqref{eq:Jsub}]. With the sub-map relaxation, however, we can adjust the sC-SKF to given hardware constraints by adjusting the number of sub-maps employed. Increasing the amount of sub-maps decreases the size of the system when solving for $\mathbf{J}_1$, dramatically reducing computation time. It should be noted, however, that increasing the number of sub-maps decreases the information associated with the map, which typically leads to slightly less accurate device pose estimates (Sect.~\ref{sec:exp}).
\section{2D-3D Feature Correspondence Pipeline} \label{corr_pipeline}
Before employing map-based updates, we identify correspondences between 2D feature measurements (FREAK features~\cite{FREAK}) in the current image and 3D features which have been previously mapped (Fig.~\ref{fig:scheme} Box B). Our pipeline takes a dual layer approach: the first considers the case when the estimator has no prior for the 4 d.o.f device-to-map transformation, while the second takes advantage of such a prior to improve efficiency.

\subsection{Pose-less correspondence generation}

 \textbf{1) VT query:} We query the saved VT with the current image, which returns up to five mapped images of similar appearance.
 
 \textbf{2) Feature matching:} We apply binary descriptor matching between features in the query and returned images.
 
 \textbf{3) Outlier rejection:} We apply 3+1 pt RANSAC~\cite{Naroditsky2012} on each image returned by the VT. If less than 7 correspondences remain, we classify that VT return as an outlier. Then, the 2+1 pt RANSAC~\cite{kukelova2011} is applied over the remaining set of inlier correspondences, If less than 13 correspondences remain, we classify all feature measurements as outliers.

\subsection{Pose-assisted correspondence generation}
During nominal operation, the estimator will have a reliable estimate for the device-map transformation. We leverage this estimate to re-project a subset of mapped features into the current image bypassing the VT (the main bottleneck in the pose-less pipeline).

 \textbf{1) Pose-based matching:} We find mapped images whose camera poses are nearby the current camera pose. A number of heuristics could be used to define nearby mapped images; in our case we require images to be within 3 m of the current position, and have an optical axis within 45 deg of the current optical axis. Images satisfying these criteria are the initial set of mapped image matches.
 
 \textbf{2) Co-visibility:} We add mapped images which view at least 50 common features with any of the pose-based image matches to the set of mapped image matches. This step allows the pipeline to include images that are far from the current camera pose, but still view the same scene.
 
 \textbf{3) Feature re-projection and matching:} The union of all features in the set of matched mapped images (through pose-based matching and the co-visibility) are reprojected onto the current image. Re-projected features are matched with current features by binary descriptor matching; projected features considered for matching are limited to a small radius (30 pixels) around the current feature.
 
 \textbf{4) Outlier rejection} We apply 2+1 pt RANSAC on the set of matched 2D-3D correspondences to reject outliers. If less than 13 correspondences remain, we classify all feature measurements as outliers.
\subsection{Additional outlier rejection}
While RANSAC-based approaches generally remove the majority of any 2D-3D outliers, we further improve our system's reliability by applying a Mahalanobis distance test on a per-feature basis.
\section{Experiment Results} \label{sec:exp}
All experimental results are obtained on a Google Project Tango tablet, which is equipped with a fisheye, global-shutter, grayscale camera, a MEMS quality IMU, a quad-core, 2.3 GHz ARM Cortex-A15 CPU, and 4 GB of RAM.

\subsection{Data collection and ground truth}
To generate experimental results we collected three datasets: the first (DS1) (2,069 images, approximately 60m long) is the dataset which the tested estimators run, the second (DS2) (2,627 images, approximately 75m long) is used to generate the map of the area of interest. All maps and sub-maps are generated using the CM method of \cite{GUO_CM_LINE}.
In addition to each map's estimate and Cholesky factor, we save FREAK binary descriptors~\cite{FREAK} and a VT~\cite{Nister062}, which indexes all mapped images. To acquire ground truth for the trajectory in DS1, we generate a BLS estimate, where in addition to all camera and IMU measurements, we supply the BLS estimator with absolute pose measurements from a VICON system.
\subsection{Results}
\begin{figure}
\centering
 \includegraphics[width=0.43\textwidth]{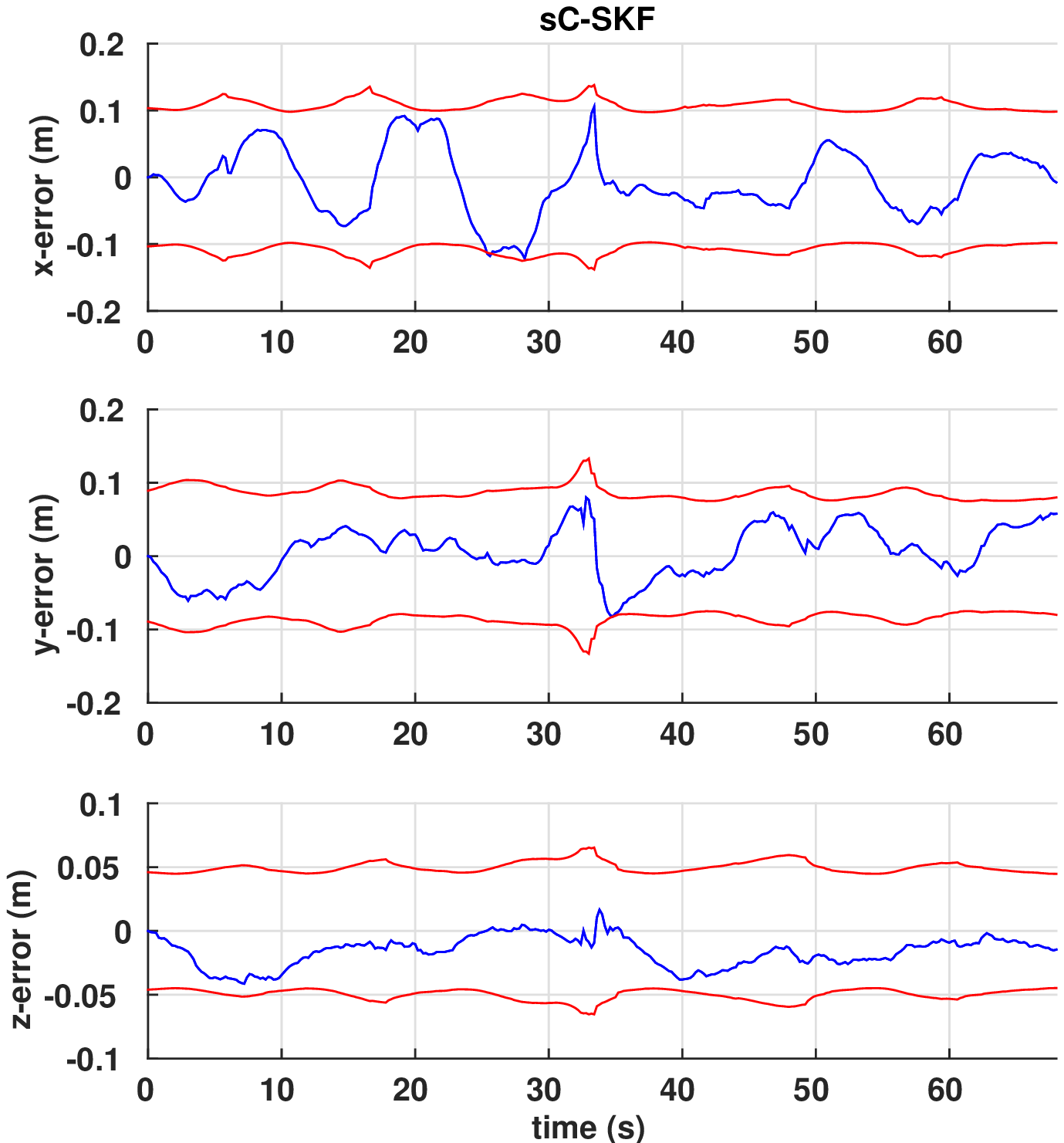}
 \includegraphics[width=0.43\textwidth]{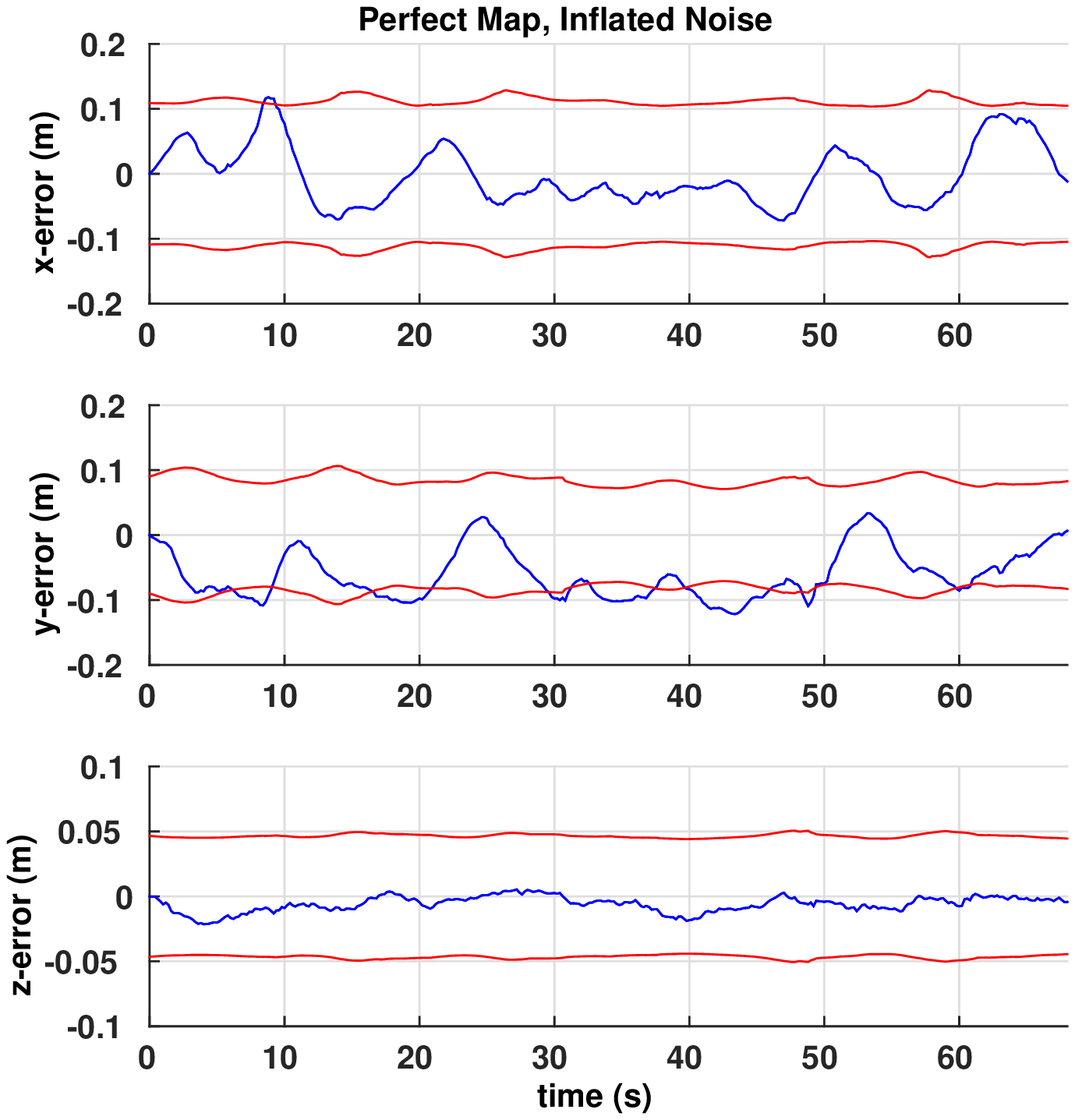}
 \caption{The error and 3$\sigma$ bounds for the sC-SKF with 2 sub-maps and perfect map-based updates with inflated measurement noise.}
 \label{fig:consistancy}
\end{figure}

Note that in the trajectory of DS1, the user moves within the VICON-room, visiting the same scene several times.
Such a trajectory will emphasize the detrimental effects of an inconsistent estimator which does not track device-map cross-correlations (the device estimate becomes strongly correlated with the map when viewing the same features multiple times). This inconsistency is illustrated in Fig.~\ref{fig:consistancy}, which shows the error and 3$\sigma$ bounds for the inflated measurement noise model ($\sigma$ = 7.5 pixels), and the sC-SKF estimator employing 2 sub-maps. Furthermore, the pose RMSE for each method of map-based updates can be found in Table~\ref{tab:RMSE}, where, as expected, the C-SKF and sC-KF outperform inconsistent methods.

The storage requirements of the map's uncertainty information is the main limitation of the SKF addressed by the C-SKF. The sizes of the Cholesky factors used for the map of DS2, its two sub-maps, as well as an additional, larger map (DS3), are available in Table~\ref{tab:mapsize}. Furthermore, we provide the corresponding size of each map's covariance. As expected, the sparse Cholesky factor requires much less disk space than the dense covariance, and grows at a slower rate as the map size increases.

\begin{table}[ht]
\resizebox{0.48\textwidth}{!}{%
\centering
 \begin{tabular}{|c |c c c c |}
 \hline
  Map ID & DS2 sub-map 1 & DS2 sub-map 2 & DS2 Full Map & DS3 \\
 \hline
  Num Feat.                &  1,838    & 1,424   &  2,203  & 12,164    \\
  Map Dim                  &  11,517   & 8,774   &  20,537 &  63,692  \\
  Size ($\mathbf{G}$)      &  45 MB    & 18.1 MB &  76 MB  & 457 MB  \\
  Size ($\mathbf{P}_{MM}$) &  530 MB   & 308 MB  &  1.2 GB & 16.2 GB \\
  \hline
 \end{tabular}}
 \caption{Sizes of various maps as well as the associated memory requirements of storing their Cholesky factor, $\mathbf{G}$, and covariance $\mathbf{P}_{MM}$.}
  \label{tab:mapsize}
\end{table}

\begin {table} 
\centering
 \begin{tabular} {|c c|}
 \hline
  Method                     & Position RMSE (cm) \\ \hline
  C-SKF, Single Map          & 6.2 \\
  sC-SKF, Two Sub-maps       & 6.6 \\
  Perfect Map Approximation  & 8.3 \\
  No Map-based Updates       & 14.7 \\
  \hline
 \end{tabular}
 \caption{Position RMSE} \label{tab:RMSE}
\end {table}

\begin{table}
\centering
\begin{tabular}{|c c|}
 \hline
   Number of Features in Map & Time to Compute $\mathbf{J}$ \\ \hline
    3,926 & 6.7 ms \\
    6,341 & 24.5 ms \\
    12,164 & 233.3 ms\\
   \hline
 \end{tabular}
 \caption{Average time to compute $\mathbf{J}$ [see ~\eqref{eq:Jdef}] for a single feature measurement in various maps.}
 \label{tab:backsolve}
\end{table}

\begin{table}
\resizebox{0.48\textwidth}{!}{%
 \begin{tabular}{|c c|}
 \hline
  Item & Mean Time (ms) \\ \hline \hline
  Mapped-feature Update (sC-SKF) & 180 \\
  Mapped-feature Update (perfect map) & 7 \\ \hline 
  Correspondence Detection (pose-less) & 52 \\
  Correspondence Detection (pose-assisted) & 36 \\  \hline 
  Harris + KLT Tracking & 20 \\
  Propagation + Local-feature Update & 34 \\
  \hline
 \end{tabular}}
 \caption{Mean times for the steps of the estimator pipeline.} \label{tab:times}
\end{table}

As mentioned earlier, the motivation for partitioning the map into sub-maps is to reduce the time required to compute $\mathbf{J}$ [see \eqref{eq:Jdef}].
We supply the average time for the back-solve required to compute $\mathbf{J}$ on a Project Tango tablet, for different map sizes in Table~\ref{tab:backsolve}. For large maps, it becomes impossible to perform real-time map-based updates with the C-SKF. On the other hand, by reducing the map size (i.e., increase number of sub-maps) and employing the sC-SKF, we significantly reduce the computation requirements.

Finally, we present the times of the most computationally-demanding-components of the sC-SKF pipeline (using the two sub-maps of DS2) in Table~\ref{tab:times}. Specifically, the components timed are the 2D-3D correspondence detection pipeline, the map-based update, the local feature tracking pipeline (Harris and KLT), and the MSCKF update. We also provide the time for mapped-feature updates using the perfect-map assumption (inflate noise) for comparison.
Our MSCKF runs at approximately 6Hz (depending on the user's motion). By summing the MSCKF update time and feature-tracking time listed, our local-measurements estimation requires on average 324 ms of CPU time per second, which leaves 676 ms to be devoted to map-based updates. The sC-SKF map-based update pipeline (2D-3D pose-based correspondence generation and map-based-update) takes 214 ms per iteration. Therefore, we can generally apply map-based updates while maintaining real-time operation. If the application or device has stricter processing requirements, we have the option to increase the number of sub-maps (decreasing the total information attributed to the map).

\section{Conclusion} \label{sec:conclusion}
In this paper, we focused on the problem of performing approximate, but consistent map-based localization.
Specifically, and motivated from the linear (in the map's size) processing cost, but quadratic memory requirements of the Schmidt-Kalman filter (SKF) when applied to map-based localization, we introduced the Cholesky (C)-SKF, which uses the map's Cholesky factor to model the information (and thus uncertainty) in the prior map.
By doing so, and given the sparsity of the Cholesky factor, the C-SKF has only linear, in the map's size, memory requirements.
Moreover, its equations are factored in such a form so as to avoid inverting the Cholesky factor of the map's Hessian matrix.
Despite, however, the gains in efficiency, the processing cost of the C-SKF may grow more than linearly in the map's size.
In order to bound its processing cost, we introduced a relaxation, termed the sC-SKF, which uses the sub-maps obtained by partitioning the original map, with minimal loss in accuracy.
Lastly, the computational requirements of the proposed C-SKF and sC-SKF were assessed using a Project Tango tablet, while we demonstrated their superior performance against other approximate, but inconsistent, map-based approaches through real-world experiments using high accuracy ground truth.  

\begin{appendices}
\section{Proof of sub-map consistency}\label{cm_appen}
We seek to prove Theorem~\ref{th} in Section~\ref{sec:sub-maps-def}.
\begin{IEEEproof}
 As described in~\cite{GUO_CM_LINE}, for the case of two sub-maps, the KKT matrix, $\mathcal{K}$, resulting from the constrained optimization problem is:
 \begin{equation*}
  \mathcal{K} = \begin{bmatrix} \mathcal{H}_1 & \mathbf{0} & \mathbf{A}^T_1 & \mathbf{0} \\
                                \mathbf{0} & \mathcal{H}_2 & \mathbf{A}^T_2 & \mathbf{0} \\
                                \mathbf{A}_1 & \mathbf{A}_2 & \mathbf{0} & \mathbf{B} \\
                                \mathbf{0} & \mathbf{0} & \mathbf{B}^T & \mathbf{0} \end{bmatrix}
 \end{equation*}
 where $\mathcal{H}_i$ is the Hessian of sub-map $i$, $\mathbf{A}_i$ corresponds to features common to the two sub-maps, and $\mathbf{B}$ is a tall matrix whose columns correspond to the 4 d.o.f transformation between the two sub-maps.
 
 Computing the covariance of the sub-maps in the CM result requires computing the 2 $\times$ 2 block sub-matrix of the inverse of $\mathcal{K}$. i.e.,
 \begin{align*}
  \mathbf{P} &= (\mathcal{K}^{-1})_{1:2, 1:2} \\
             &= \left\{ \begin{bmatrix} \mathcal{H}_1 & \mathbf{0} \\
                                                                       \mathbf{0} & \mathcal{H}_2 \end{bmatrix} - \begin{bmatrix} \mathbf{A}^T_1 & \mathbf{0} \\
                                                                       \mathbf{A}^T_2 & \mathbf{0} \end{bmatrix} \begin{bmatrix} \mathbf{0} & \mathbf{B} \\
                                                                       \mathbf{B}^T & \mathbf{0} \end{bmatrix}^{-1} \begin{bmatrix} \mathbf{A}_1 & \mathbf{A}_2 \\
                                                                       \mathbf{0} & \mathbf{0} \end{bmatrix}\right\}^{-1}
 \end{align*}
 Employing the matrix inversion lemma yields:
 \begin{equation*}
  \mathbf{P} = \begin{bmatrix} \mathcal{H}^{-1}_1 & \mathbf{0} \\
                               \mathbf{0} & \mathcal{H}^{-1}_2 \end{bmatrix} - 
                \begin{bmatrix} \mathcal{H}^{-1}_1 & \mathbf{0} \\
                               \mathbf{0} & \mathcal{H}^{-1}_2 \end{bmatrix} \mathbf{M}
                  \begin{bmatrix} \mathcal{H}^{-1}_1 & \mathbf{0} \\
                               \mathbf{0} & \mathcal{H}^{-1}_2 \end{bmatrix}
 \end {equation*}
 where
 \begin{align*}
  \mathbf{M} = \begin{bmatrix} \mathbf{A}^T_1 & \mathbf{0} \\
                                \mathbf{A}^T_2 & \mathbf{0} \end{bmatrix}
                \mathbf{W}
                                \begin{bmatrix} \mathbf{A}_1 & \mathbf{A}_2 \\
                                 \mathbf{0} & \mathbf{0} \end{bmatrix}  \label{eq:CM_M} \numberthis
 \end{align*}
 and
 \begin{align*}
 \mathbf{W}^{-1} &= \begin{bmatrix} \mathbf{A}_1 & \mathbf{A}_2 \\
                                 \mathbf{0} & \mathbf{0} \end{bmatrix}
                \begin{bmatrix} \mathcal{H}^{-1}_1 & \mathbf{0} \\
                               \mathbf{0} & \mathcal{H}^{-1}_2 \end{bmatrix}
                \begin{bmatrix} \mathbf{A}^T_1 & \mathbf{0} \\
                                \mathbf{A}^T_2 & \mathbf{0} \end{bmatrix}
              -    \begin{bmatrix} \mathbf{0} & \mathbf{B} \\
                                \mathbf{B}^T & \mathbf{0} \end{bmatrix}  \\
 &= \begin{bmatrix} \bm{\Theta} & -\mathbf{B} \\ -\mathbf{B}^T & \mathbf{0} \end{bmatrix}
 \end{align*}
 with 
 \begin{align*}
   \bm{\Theta} &\triangleq \mathbf{A}_1 \mathcal{H}^{-1}_1 \mathbf{A}_1^T + \mathbf{A}_2 \mathcal{H}^{-1}_2 \mathbf{A}_2^T
 \end{align*}

 On the other hand, if we treat each of the sub-maps as being independent (i.e., we ignore the constraints imposed by common features) but we compute their Hessians using the CM result, their covariances are:
 \begin{equation}
  \mathbf{\bar{P}} = \begin{bmatrix} \mathcal{H}^{-1}_1 & \mathbf{0} \\
                               \mathbf{0} & \mathcal{H}^{-1}_2 \end{bmatrix} 
 \end{equation}
 In order to prove $\mathbf{P} \preceq \mathbf{\bar{P}}$, it suffices to show $\mathbf{M}$ in~\eqref{eq:CM_M} is symmetric positive semi-definite (PSD).
 We start by computing
 \begin{align*} \mathbf{W} = \begin{bmatrix} \mathbf{X} & \mathbf{Y} \\ \mathbf{Y}^T & \mathbf{Z} \end{bmatrix} \label{eq:xyz} \numberthis
\end{align*}
                
                where:
                \begin{align}
                 \mathbf{X} &= \bm{\Theta}^{-1} - \bm{\Theta}^{-1} \mathbf{B} \left( \mathbf{B}^T \bm{\Theta}^{-1} \mathbf{B} \right) ^{-1}   \mathbf{B}^T \bm{\Theta}^{-1} \label{eq:x}\\ 
                 \mathbf{Y} &= - \bm{\Theta}^{-1} \mathbf{B} \left( \mathbf{B}^T \bm{\Theta}^{-1} \mathbf{B} \right)^{-1} \\
                 \mathbf{Z} &= -  \left(\mathbf{B}^T \bm{\Theta}^{-1} \mathbf{B} \right)^{-1} 
                \end{align}
Substituting \eqref{eq:xyz} in \eqref{eq:CM_M} yields
\begin {equation}
\mathbf{M} =  \begin{bmatrix} \mathbf{A}^T_1 \\ \mathbf{A}^T_2 \end{bmatrix} \mathbf{X} \begin{bmatrix} \mathbf{A}_1 & \mathbf{A}_2 \end{bmatrix}
\end{equation}
Note that $\mathbf{X}$ in~\eqref{eq:x} is PSD since it is the (1,1) Schur complement of the PSD matrix $ \begin{bmatrix} \mathbf{I} \\ \mathbf{B}^T \end{bmatrix} \bm{\Theta}^{-1} \begin{bmatrix} \mathbf{I} & \mathbf{B} \end{bmatrix}$. Thus $\mathbf{M}$ is also PSD.
\end{IEEEproof}

\end{appendices}

\bibliographystyle{plainnat}
\bibliography{references}

\begin{thebibliography}{25}
\providecommand{\natexlab}[1]{#1}
\providecommand{\url}[1]{\texttt{#1}}
\expandafter\ifx\csname urlstyle\endcsname\relax
  \providecommand{\doi}[1]{doi: #1}\else
  \providecommand{\doi}{doi: \begingroup \urlstyle{rm}\Url}\fi

\bibitem[Agrawal and Konolige(2008)]{konolige08tro}
Motilal Agrawal and Kurt Konolige.
\newblock Frameslam: From bundle adjustment to real-time visual mapping.
\newblock \emph{{IEEE} Trans. on Robotics}, 24\penalty0 (5):\penalty0
  1066--1077, October 2008.

\bibitem[Alahi et~al.(2012)Alahi, Ortiz, and Vandergheynst]{FREAK}
Alexandre Alahi, Raphael Ortiz, and Pierre Vandergheynst.
\newblock {FREAK}: Fast retina keypoint.
\newblock In \emph{Proc. of the IEEE Conference on Computer Vision and Pattern
  Recognition}, pages 510--517, College Park, MD, June 16--21 2012.

\bibitem[Arth et~al.(2009)Arth, Wagner, Klopschitz, Irschara, and
  Schmalstieg]{clemens09}
Clemens Arth, Daniel Wagner, Manfred Klopschitz, Arnold Irschara, and Dieter
  Schmalstieg.
\newblock Wide area localization on mobile phones.
\newblock In \emph{8th {IEEE} and {ACM} International Symposium on Mixed and
  Augmented Reality}, pages 73--82, Orlando, FL, USA, October 19--22 2009.

\bibitem[Chatfield(1997)]{chatfield1997fundamentals}
Averil~Burton Chatfield.
\newblock \emph{{Fundamentals of High Accuracy Inertial Navigation}}, volume
  174.
\newblock American Institute of Aeronautics and Astronautics, 1997.

\bibitem[Choudhary et~al.()Choudhary, Carlone, Christensen, and
  Dellaert]{choudhary2015}
Siddharth Choudhary, Luca Carlone, Henrik~I Christensen, and Frank Dellaert.
\newblock Exactly sparse memory efficient slam using the multi-block
  alternating direction method of multipliers.
\newblock In \emph{Proc. of the {IEEE/RSJ} International Conference on
  Intelligent Robots and Systems}, Hamburg, Germany.

\bibitem[Guivant and Nebot(2001)]{guivant2001optimization}
Jose~E Guivant and Eduardo~Mario Nebot.
\newblock Optimization of the simultaneous localization and map-building
  algorithm for real-time implementation.
\newblock \emph{{IEEE} Trans. on Robotics and Automation}, 17\penalty0
  (3):\penalty0 242--257, 2001.

\bibitem[Guo et~al.(2014)Guo, Kottas, DuToit, Ahmed, Li, and
  Roumeliotis]{Guo-RSS-14}
Chao Guo, Dimitrios Kottas, Ryan DuToit, Ahmed Ahmed, Ruipeng Li, and Stergios
  Roumeliotis.
\newblock Efficient visual-inertial navigation using a rolling-shutter camera
  with inaccurate timestamps.
\newblock In \emph{Proceedings of Robotics: Science and Systems}, Berkeley, CA,
  USA, July 12--16 2014.

\bibitem[Guo et~al.(2016)Guo, Sartipi, DuToit, Georgiou, Li, O'Leary, Nerurkar,
  Hesch, and Roumeliotis]{GUO_CM_LINE}
Chao Guo, Kourosh Sartipi, Ryan DuToit, Georgios Georgiou, Ruipeng Li, John
  O'Leary, Esha Nerurkar, Joel Hesch, and Stergios Roumeliotis.
\newblock Large-scale cooperative 3{D} visual-inertial mapping in a {M}anhattan
  world.
\newblock In \emph{Proc. of the {IEEE} International Conference on Robotics and
  Automation}, Stockholm, Sweden, May 16--21 2016.
\newblock URL \url{http://mars.cs.umn.edu/papers/CM_line.pdf}.

\bibitem[Harris and Stephens(1988)]{Harris1988}
Chris Harris and Mike Stephens.
\newblock A combined corner and edge detector.
\newblock In \emph{Proc. of the Alvey Vision Conference}, pages 147--151,
  Manchester, UK, August 31 -- September 2 1988.

\bibitem[Hesch et~al.(2014)Hesch, Kottas, Bowman, and
  Roumeliotis]{Hesch_TRO_14}
Joel~A. Hesch, Dimitrios~G. Kottas, Sean~L. Bowman, and Stergios~I.
  Roumeliotis.
\newblock Consistency analysis and improvement of vision-aided inertial
  navigation.
\newblock \emph{{IEEE} Trans. on Robotics}, 30\penalty0 (1):\penalty0 158--176,
  February 2014.

\bibitem[Julier(2001)]{Julier2001}
Simon~J Julier.
\newblock A sparse weight {Kalman} filter approach to simultaneous localisation
  and map building.
\newblock In \emph{Proc. of the {IEEE/RSJ} International Conference on
  Intelligent Robots and Systems}, volume~3, pages 1251--1256, Maui, HI, USA,
  October 29 -- November 3 2001.

\bibitem[Klein and Murray(2007)]{klein2007parallel}
Georg Klein and David Murray.
\newblock Parallel tracking and mapping for small {AR} workspaces.
\newblock In \emph{6th {IEEE} and {ACM} International Symposium on Mixed and
  Augmented Reality}, pages 225--234, Nara, Japan, November 13--16 2007.

\bibitem[Kukelova et~al.(2011)Kukelova, Bujnak, and Pajdla]{kukelova2011}
Zuzana Kukelova, Martin Bujnak, and Tomas Pajdla.
\newblock Closed-form solutions to minimal absolute pose problems with known
  vertical direction.
\newblock In \emph{Proc. of the Asian Conference on Computer Vision}, pages
  216--229, Queenstown, New Zealand, November 8--12 2011.

\bibitem[Lucas and Kanade(1981)]{Kanade1981}
Bruce~D. Lucas and Takeo Kanade.
\newblock An iterative image registration technique with an application to
  stereo vision.
\newblock In \emph{Proc. of the International Joint Conference on Artificaial
  Intelligence}, pages 674--679, Vancouver, British Columbia, August 24--28
  1981.

\bibitem[Lynen et~al.(2015)Lynen, Sattler, Bosse, Hesch, Pollefeys, and
  Siegwart]{Lynen2015}
Simon Lynen, Torsten Sattler, Michael Bosse, Joel Hesch, Marc Pollefeys, and
  Roland Siegwart.
\newblock Get out of my lab: Large-scale, real-time visual-inertial
  localization.
\newblock In \emph{Proc. of Robotics: Science and Systems Conference}, Rome,
  Italy, July 13--17 2015.

\bibitem[Middelberg et~al.(2014)Middelberg, Sattler, Untzelmann, and
  Kobbelt]{Middelberg2014}
Sven Middelberg, Torsten Sattler, Ole Untzelmann, and Leif Kobbelt.
\newblock Scalable 6-dof localization on mobile devices.
\newblock In \emph{Proc. of the European Conference on Computer Vision}, pages
  649--663, Zurich, Switzerland, September 6--12 2014.

\bibitem[Mirzaei and Roumeliotis(2008)]{Mirzaei08}
Faraz~M. Mirzaei and Stergios~I. Roumeliotis.
\newblock A {Kalman} filter-based algorithm for {IMU}-camera calibration:
  Observability analysis and performance evaluation.
\newblock \emph{{IEEE} Trans. on Robotics}, 24\penalty0 (5):\penalty0
  1143--1156, October 2008.

\bibitem[Mourikis et~al.(2009)Mourikis, Trawny, Roumeliotis, Johson, Ansar, and
  Matthies]{Mourikis09}
Anastasios~I. Mourikis, Nikolas Trawny, Stergios~I. Roumeliotis, Andrew~E.
  Johson, Adnan Ansar, and Larry Matthies.
\newblock Vision-aided inertial navigation for spacecraft entry, descent, and
  landing.
\newblock \emph{{IEEE} Trans. on Robotics}, 25\penalty0 (2):\penalty0 264--280,
  April 2009.

\bibitem[Naroditsky et~al.(2012)Naroditsky, Zhou, Gallier, Roumeliotis, and
  Daniilidis]{Naroditsky2012}
Oleg Naroditsky, Xun~S. Zhou, Jean Gallier, Stergios~I. Roumeliotis, and Kostas
  Daniilidis.
\newblock Two efficient solutions for visual odometry using directional
  correspondence.
\newblock \emph{{IEEE} Trans. on Pattern Analysis and Machine Intelligence},
  34\penalty0 (4):\penalty0 818--824, April 2012.

\bibitem[Nister and Stewenius(2006)]{Nister062}
David Nister and Henrik Stewenius.
\newblock Scalable recognition with a vocabulary tree.
\newblock In \emph{Proc. of the {IEEE} Conference on Computer Vision and
  Pattern Recognition}, pages 2161--2168, New York, NY, June 17--22 2006.

\bibitem[Rublee et~al.(2011)Rublee, Rabaud, Konolige, and Bradski]{ORB}
Ethan Rublee, Vincent Rabaud, Kurt Konolige, and Gary Bradski.
\newblock {ORB}: {A}n efficient alternative to {SIFT} or {SURF}.
\newblock In \emph{Proc. of the {IEEE} International Conference on Computer
  Vision}, pages 2564--2571, Barcelona, Spain, November 6--13 2011.

\bibitem[Schmidt(1966)]{schmidt1966}
Stanley~F. Schmidt.
\newblock Applications of state space methods to navigation problems.
\newblock \emph{in C. T. Leondes, Editor, Advanced Control Systems},
  3:\penalty0 293--340, 1966.

\bibitem[Simon(2006)]{Simon-textbook}
Dan Simon.
\newblock \emph{Optimal state estimation: {Kalman}, H infinity, and nonlinear
  approaches}.
\newblock John Wiley \& Sons, 2006.

\bibitem[Triggs et~al.(2000)Triggs, McLauchlan, Hartley, and
  Fitzgibbon]{Triggs00}
Bill Triggs, Philip McLauchlan, Richard Hartley, and A.~Fitzgibbon.
\newblock Bundle adjustment - a modern synthesis.
\newblock In \emph{Vision Algorithms: Theory and Practice}, pages 298--375.
  Springer--Verlag, 2000.

\bibitem[Ventura et~al.(2014)Ventura, Arth, Reitmayr, and
  Schmalstieg]{Ventura2014}
Jordi Ventura, Clemens Arth, Gerhard Reitmayr, and Dieter Schmalstieg.
\newblock Global localization from monocular {SLAM} on a mobile phone.
\newblock \emph{{IEEE} Trans. on Visualization and Computer Graphics},
  20\penalty0 (4):\penalty0 531--539, April 2014.

\end{thebibliography}

\end{document}